# Structured Bayesian Compression for Deep models in mobile-enabled devices for connected healthcare

Sijia Chen, Bin Song, *Member, IEEE,* Xiaojiang Du, *Senior Member, IEEE, Nadra Guizani, Fellow, IEEE*

*Abstract*—Deep Models, typically Deep neural networks, have millions of parameters, analyze medical data accurately, yet in a time-consuming method. However, energy cost effectiveness and computational efficiency are important for prerequisites developing and deploying mobile-enabled devices, the mainstream trend in connected healthcare. Therefore, deep models' compression has become a problem of great significance for real-time health services. In this article, we first emphasize the use of Bayesian learning for model sparsity, effectively reducing the number of parameters while maintaining model performance. Specifically, with sparsity inducing priors, large parts of the network can be pruned with a simple retraining of arbitrary datasets. Then, we propose a novel structured Bayesian compression architecture by adaptively learning both group sparse and block sparse while also designing sparse-oriented mixture priors to improve the expandability of the compression model. Experimental results from both simulated datasets (MNIST) as well as practical medical datasets (Histopathologic Cancer) demonstrate the effectiveness and good performance of our architecture on Deep model compression.

*Index Terms*—Deep model compression, Structured Bayesian compression framework, group sparse, block sparse, mixture priors

## I. Introduction

WITH the revolutionary innovation and breakthroughs in the deep model, models based on deep neural networks (DNN) have been utilized for various tasks, such as image classification, detection, and object segmentation, (e.g. ). This type of model has been introduced into the healthcare field as an aid to disease diagnosis because of its proven high reliability and accuracy [1]. To achieve effective learning of medical data, there are always millions of parameters with complex and dense network architecture. Currently, the node devices of connected healthcare have been widely extended to mobile-enabled devices, such as phones, portable medical devices and small service stations [2]. However, those deep models cannot be transferred to such scenarios due to limitations of mobile-enabled devices, such as storage space, available computing units, and real-time requirements. Therefore, efficiency-oriented model compression is significantly crucial for mobile-based services in connected healthcare. Several papers (e.g. [12], [13], [14]) have studied related mobile and wireless issues.

Although the optimized trade-off between compression rate and performance cannot always be maintained, the major motivation for this is that the majority capabilities of DNN-based models come from only a small amount of structures and parameters [3].

Compressing the model from a Bayesian point of view with sparsity inducing priors leads to an efficient sparse method that aims to promote the distribution of the parameter to zero with extremely thin tail. Therefore, majority parameters can be adaptively neglected during the training. That is, instead of introducing complex processes after learning, Bayesian compression can control the training process through specific priors so that the model can automatically learn posterior distributions, essentially guiding the sparseness of model structures and parameters.

However, Bayesian-based compression methods have some limitations, including the selection of priors and potential structure detection. Table 1 shows a comparison of sparsity inducing distributions. The horseshow, normal-Jeffreys priors maintain same heave tails, similar to Cauchy, which decays with $\theta^2$ and has an infinitely tall spike near zero. Also, double-exponential has a lighter tail compared to all other distributions, making it a priority choice for keeping weights with small values. In contrast, the horseshoe prior maintains enough probability mass for the in-between values of $\lambda$ and thus can potentially offer better regularization and generalization. Therefore, selecting a suitable prior for a Bayesian compression is a hard task. Furthermore, existing Bayesian-based solutions concentrate on each weight without learning the structure of the

This work was supported by the National Natural Science Foundation of China under Grant (No. 61772387 and No. 61802296), Fundamental Research Funds of Ministry of Education and China Mobile (MCM20170202), China Postdoctoral Science Foundation Grant (No. 2017M620438), the Fundamental Research Funds for the Central Universities, and supported by ISN State Key Laboratory.

Sijia Chen is working in the State Key Laboratory of Integrated Services Networks, Xidian University, Xi'an 710071, China (e-mail: sjchen@stu.xidian.edu.cn).

Bin Song is the corresponding author who is working in the State Key Laboratory of Integrated Services Networks, Xidian University, Xi'an 710071, China (e-mail: bsong@mail.xidian.edu.cn).

Xiaojiang Du is working in Dept of Computer and Information Sciences, Temple University, Philadelphia, PA, USA, (Email: dxj@ieee.org).

Nadra Guizani is working for Dept. of Electrical and Computer Engineering, University of Idaho, Moscow, Idaho, USA, (email: nguizani@ieee.org)

DNN-based model. For instance, Fig. 1 illustrates both the hierarchical structure among layers as well as the correlation between potential weights within one layer. Learning these structures existing in the model will encourage to the group sparse, accelerating learning convergence and increasing the compression ratio of homogeneous network nodes. In this paper, we propose a novel structured Bayesian compression model which mainly contains the following two principles.

**Mixture of sparsity inducing prior**: To avoid the possibility of mis-specifying the prior by using prior incorrect knowledge, we assign the prior to weights by finite mixtures of natural conjugate sparsity inducing priors. These priors tend to approximate an optimal prior that best fits the compression process and data. By considering prior mixing weights as learnable hyperparameter that can be estimated from a Bayesian perspective, flexibility and complementarity of these distributions can be introduced to our model.

**Structured Sparsity Learning**: Weights in different layers and the weights partitioned within one layer are two typical structures in DNN-based model. The former can be directly determined when the network structure is known, whereas the latter is often unknown. Therefore, we propose a Structured Sparsity Learning method to both utilize the intrinsic structure and to learn potential block structure within weights in one layer.

The contribution of our paper exists in three parts. Firstly, we introduce the Bayesian-based model compression method into the application of mobile-enabled devices in connected healthcare, which can significantly extend the available fields of DNN-based models in healthcare as well as improve the flexibility of compression. Secondly, a learnable mixture sparsity inducing priors mechanism is applied to the sparsity Bayesian learning to improve the flexibility of compression. Thirdly, the inherent hierarchical structure and the potential relationships that exist between the parameters within the layer are used for DNN-based model compression. Overall, we propose a novel structured Bayesian compression architecture for DNN-based models used in mobile devices that have limited hardware resources in connected healthcare.

The rest of this article is arranged as follows. In the next section, we describe the related work in model compression, especially Bayesian compression. Then, we present our structured Bayesian compression architecture following the experimental results on both synthetic dataset and practical medical dataset. Finally, the conclusion is presented.

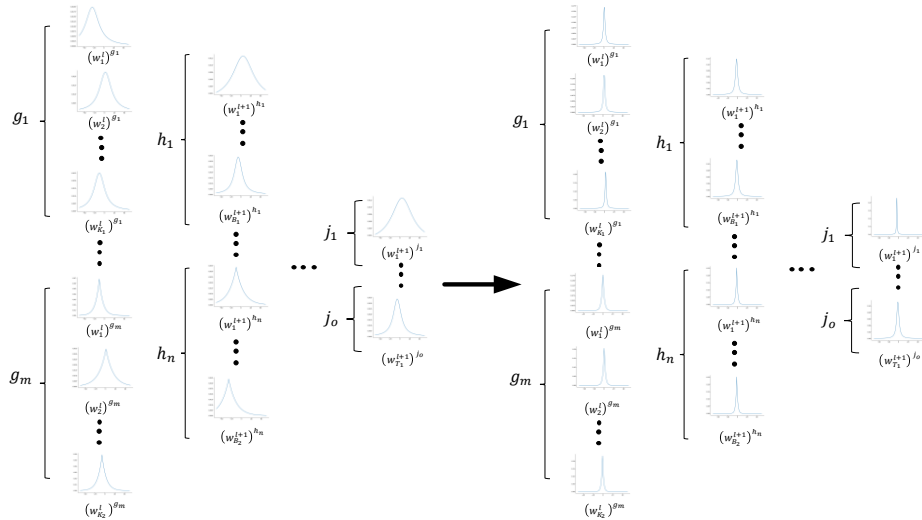

Fig. 1. The illustration of the model compression. We present the common idea of the Bayesian learning method used to compress weights in DNN-based model.

TABLE I
COMPARISON OF DIFFERENT SPARSITY INDUCING PRIORS

| Sparsity inducing distributions | Probability density | | shortage |
|---|---|---|---|
| | Near 0 | Tail | |
| Cauchy | | | Infinite spikes at zero |
| Laplace | | | Not provide uncertainty |
| Horseshoe | | | Relatively heavy tailed |
| Normal-Jeffreys(log-uniform) | | | Non-normalizable |
| Spike-and-slab | | | Computationally expensive inference |

## II. RELATED WORK

The mainstream compression methodologies for the DNN-based model derive from two perspectives, including model storage optimization as well as sparseness.

### A. Optimization

For optimization methods, the main methods include pruning, trained quantization, and encoding. Specifically, since the majority connections and weights in the DNN-based model is redundance and only provides minor contributions, the network can be pruned by learning only the important connections [5]. Because of the correlations and similarities that exist between magnitude of weights, parameters can be quantized to introduce a shared mechanism to store specific values instead of all weights [6]. Then, the required storage space for weights can be further reduced by using variable-length codewords (Huffman coding methods) to encode source symbols. In this type of methodology, the main principle is fine tuning, which involves training the learned network. [4] combines these ideas together to form a three-stage pipeline: pruning, trained quantization, and Huffman coding. It constructs a framework to reduce the storage requirement of neural networks by 35× to 49× without affecting models' accuracy. It had been demonstrated that these methods can remove the over-parametrized and significant redundancy for deep learning models, saving both computation and memory usage.

### B. Bayesian Compression

The main disadvantages of optimization are tedious adjustment process and the over-fitting of parameters and data sets, making it hard to transfer to various applications and tasks efficiently. The idea of sparseness can avoid these issues while achieving accuracy-efficiency balance and taking the matrix format of weights into consideration.

In recently years, there have been several encouraging results obtained by Bayesian approaches that employ model compression [7], [8], [10], [11]. In these methods, authors tend to achieve the Compressed Sparse Column (CSC) format and Optimal NN architecture from the Bayesian viewpoint. That is, by introducing sparsity inducing priors, most of the redundant weights can learn a posterior distribution that shrinks to zero or has higher uncertainties. This encourages an efficiency aware architecture, and optimization methods can be used to further compress the model afterwards.

Unlike optimized compression algorithms that require many cumbersome steps, Bayesian compression is a way to tackle efficiency and compression in DNN-based models in a unified and theoretically principled way. Overall, Bayesian compression can improve energy efficiency and obtain real-time models with reduced computation while maintaining the performance of the models.

## III. OVERVIEW OF THE STRUCTURED BAYESIAN COMPRESSION FRAMEWORK

Following the idea of sparse learning based on Bayesian learning, a Bayesian sparse learning method constructed with mixture sparsity inducing distributions as well as a learnable structure group sparsity algorithm is proposed to encourage flexibility, a higher percentage of parameter shrinkage, and faster convergence. In Fig. 2, we present an overview of the proposed framework with each component, from prior mixture design to structured learning, each of which is described below.

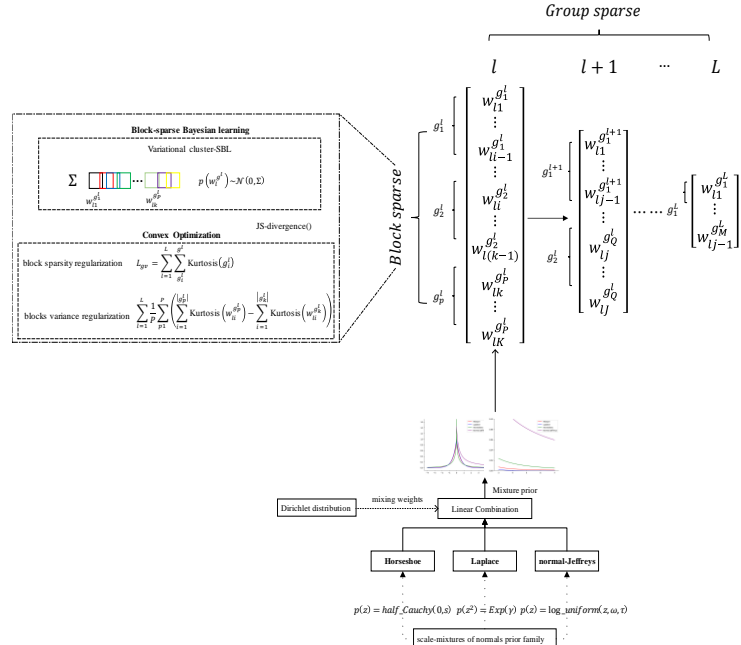

Fig. 2. The overall architecture of our proposed structured Bayesian compression framework. In the pictures contained in this figure, we compare the combinations of prior with each of its components.

### A. Mixture of sparsity inducing priors

The selection of this Bayesian prior model is a critical issue in our sparsity-oriented framework. As shown in Table 1, those priors present different sparsity motivations around 0 and at the tail, as well as in the shrink curve. Mixture distributions which can be represented as a linear superpositions are widely used as computationally convenient representations for modelling multitask frameworks. After combination, the combined distribution can be used as the prior of Bayesian learning, the posterior of which can be inferred by variational learning or Markov chain Monte Carlo sampling.

On one hand, mixture distributions can protect designers from substantial but incorrect prior knowledge. That is to say that the use of mixture priors eliminates the effect of mis-specified prior. On the other hand, mixture components specify a task's information and probability distribution characteristics. They may not have optimal meaning, but can complement each other to increase the range of learnability. Also, mixture models are frequently referred to as semi-parametric models since their flexibility allows them to approximate non-parametric problems and their ability to describe complex behavior in the learning process.

The determination of component distributions and suitable prior mixing weights of components are two key tasks in the application of mixture distributions. In most cases, a pre-defined Dirichlet distribution is used as the top combination mechanism, in which representative mixing coefficients softly mix a list of distribution objects.

A widely used family of distributions is known as scale-mixtures of normals, which are quite general and behave as the source of many well-known sparsity inducing distributions. That is, a parameter w is present in a zero-mean normal distribution N(w;0, $z^2$) whose variance z is governed by a distribution p(z). If z has heavier tails and many of its entries are zero, or nearly so, the marginal prior distributions of a random variable w can be recovered over it such that the biases in the posterior distribution over w can be sparse. For the consistency of component distribution, we adopt scale-mixtures of normals prior family as component distribution to form a mixture prior to achieve compression and efficiency in neural networks. In this paper, we consider three distributions as follows.

**Horseshoe distribution**: The horseshoe prior is both robust and free of manual hyperparameters while maintaining high performance in sparse situations. Compared with other local shrinkage rules, it ensures that the model converges to a reasonable location with a highly efficient rate. By combining strong global shrinkage and local adaptation, it creates a good procedure for sparsity by the shrinkage rule. Therefore, the horseshoe is an attractive default choice among shrinkage priors. The specific distribution is $(w_{ij}|z_i) \sim \mathcal{N}(0, z_i^2)$, $(z_i|\sigma) \sim C^+(0, \tau)$, $\tau \sim C^+(0, \sigma)$ where $\sigma$ is the parameter that can be learned.

**Laplace distribution:** It is a continuous probability distribution which is also known as a double exponential distribution. Within scale-mixtures of a normal family, we can get the Laplace distribution by setting $p(z^2) = e^\lambda$. Computationally, neural networks can be pruned by setting a Laplace prior because of the corresponding posterior distribution of the Lasso estimator. However, it lacks uncertainty because point estimate cause it to over-fit.

**Normal-Jeffreys distribution**: Jeffreys priors are limits of conjugate prior densities, which are widely used in Bayesian analysis for single parameter models. After marginalizing over the scales z with $p(z) \propto |z|^{-1}$, we can introduce w as a Jeffreys prior. This prior has a very strong shrinkage probability at zero but also has the an extremely heavy tail which makes it non-normalizable.

The rationale of the prior is to benefit from the sparse ability at zero as well as to avoid its heavy tail tendency by combining the above distributions with a smoothness promoting distribution. After analyzing contents shown both in Fig. 3 and in Table 1, in this paper, we utilize finite mixtures which involve the combination of the above distributions to form the prior of the framework.

Furthermore, to get the posterior distribution in a fully Bayesian treatment, variational learning is utilized to achieve the inference. In a Bayesian inference, we firstly regard prior mixing weights as a Dirichlet distribution. The parameters α are modelled by hyperparameters that should be learned during the training process. Following the proposed procedure in [8] as well as the variational inference in [7], we can train the whole DNN-based model to obtain sparsity weights.

### B. Structured Sparsity Learning

As the DNN-based model presents a significant partition structure, we can exploit the structure of neural networks for our Bayesian framework to obtain better compression performance. Specifically, hierarchical network structure is an inherent way to divide weights that belong to a specific layer into different groups, shown in Fig. 1. That is, we can regard weights in each layer as correlated random variables with same shrinkage rate and behaviours which can be simultaneously modelled. Also, within each layer, weights that share the same importance before training can be automatically aggregated into clusters which contain weights with similar functionality and importance within clusters while maintaining differences in these aspects among clusters due to different connected neutral locations. Therefore, the parameters within each layer have a potential block structure. As shown in Fig. 2, elements in weight vectors belong to different clusters, causing that the entire vector to essentially be a splicing of multiple sub-blocks. As a result, given two types of structures, we can utilize them to propose a structured compression algorithm. In this way, both compression efficiency and the ratio can be significantly improved.

Therefore, we propose a structured Bayesian compression method to exploit the above structural information in DNN-based models. As for the hierarchical network structure, including layers and feature maps, we share the scales of each component in the mixture prior distribution in each group to couple the weights within the group. This leads to group sparsity. The implementation of group sparsity can be achieved by following the methods described in [7] but with a complex variational inference algorithm which will be described below.

As for the potential block structure within each layer or feature map which is illustrated in Fig. 2, both the number of blocks as well as the elements of each block should be defined. However, in our compression model, the exact block partition of weights is not available, because determining this information is an NP hard problem. Also, considering the Bayesian inference learning with sparsity inducing priors, only a few blocks contain weights with a widely distributed non-zero distribution among all blocks. This encourages weights in each layer to be block-sparse signals. Therefore, we can define the Bayesian sparse learning under these conditions as a generalized block sparse learning with the unknown block partition. This requires minimum priori knowledge on block structure. Following this weaker structure assumption, we only assume that the size of each blocks is the same so that we can avoid a strong structure learning. Besides, to improve flexibility and generality, we relax the location of the blocks. That is, each block can be arbitrarily located with an appropriate overlap such that the size of blocks is essentially different. Given this definition, we can achieve a block sparse Bayesian learning. Furthermore, it has been proven that the inter-block correlation based on block-coherence measurement and the intra-block correlation based on entry-correlation measurement can further improve the performance of model.

To implement the above structured compression algorithm, we introduce a hierarchical variational inference method. For the mixture prior, variational inference for Dirichlet process (DP) mixtures proposed in [8] is utilized to obtain an effective solution. However, this method is proposed for DP mixtures of Gaussians, whereas the component of our mixture prior belongs to the scale-mixtures of the normals prior family. Therefore, we extend this algorithm to combine various distributions that still belong to the same family by introducing a compound variational algorithm which can infer posterior for the mixture prior in a top-bottom mode with the factorized approximation. For group sparsity inducted by the hierarchical structure, following the proposed variational inference method, scales of weights within each layer are shared. For block sparsity inducted by a potential block structure within each layer or feature map, based on the proposed method Cluster-SBL in [9], we achieve the cluster learning via convex optimization. That is, under a sparsity constraint both on weights and the coefficients of clusters, we design the two as regularization in the objective function. Also, to maintain the expressive ability of the layer constructed by blocks, the block variance regularization is also added to force the distribution of the grouped weights to be skewed. The structure of the proposed method is shown in Fig. 2.

## IV. EXPERIMENTS

Two experiments are used to illustrate the effectiveness and high performance of the proposed Structured Bayesian Compression framework. In preliminary experiments, we test its model compression capabilities using the classic synthetic dataset, MNIST. Then, we introduce the proposed framework into the practical task Histopathologic Cancer Detection (HCD), such that the real time DNN-based model that is efficient and low in computational consumption can be implemented. For both experiments, evaluation criteria are demonstrated in the corresponding section.

We describe the settings of our experiments below.

**Experiment setting**: As for benchmark architectures, we use our framework on the well-known architectures of LeNet-300-100 and LeNet5 to validate its compression capabilities. Also, we compare the obtained compression performance in typical layers with the Deep Bayesian compression method as well as their original architectures. We show the amount of neurons left after pruning along with the average bit precisions for the weights at each layer. Specifically, test error (error%) and percentage of non-zero weights as well as the compression ratio (CR) are evaluated. Furthermore, those results have the best accuracy and compression ratio points (I.e. The point that the compression ratio begins to decrease), shown in the corresponding compression-accuracy curve shown in Fig. 3. The benchmark algorithms are the Deep Compression method (DC)[4], Bayesian Compression method (BC) [7] and Variational dropout sparsifies method (SVD) [10].

**Training Setting**: As for network initialization, we simply initialize the weights with those of pretrained models that is needed to be compressed. For the experiment on the MNIST dataset, the number of samples in training sets and test sets is 60000 and 10000 respectively. The batch size is 128. However, we optimize the variational lower bound scaled by the number of data points. The method is also used in DC, which can maintain adaptive hyper-parameter settings. As for scale parameter settings in the mixture prior, we set the scale of each component to a small value which is lower than $e^{-6}$ to motivate the sparse learning. For the experiment on HCD dataset, 144000 and 16000 images with equal numbers of images come from each class, and are used for training and testing, respectively. The batch size is 10 and the image size is 96. We use Adam as the optimizer and the learning rate is 0.0001. After 20 epochs, we remove the weights that shrink to zero or contain high level of uncertainly. The structure of this architecture is $[(32,3) \times 3\text{-}(63,3) \times 3\text{-}(128,3) \times 3\text{-}256\text{-}2]$ where meanings of the number are (filter size, kernel size) × layer number.

**Evaluation Setting**: To demonstrate the performance of the model, we compare our method to benchmark algorithms in the following different evaluations: the architecture of the compressed model only by pruning; the corresponding error (Error%) and compression rate (CR) of the model's performance on test dataset when there is a significantly drop in accuracy and the proportion of parameters that equals to zero in all parameters (I.e. $\frac{|W \neq 0|}{|W|}$, donated as WR). On the second dataset HCD, we mainly illustrate the efficiency and energy saving abilities of the compressed model based on the proposed framework. Particularly, we compare the number of weights in typical layers before and after compression, while testing the time needed for evaluation on a test dataset (16000 images) in a single CPU environment.

Firstly, we present the sparsity capabilities of the proposed framework by comparing the number of weights in the Pruned column in Table II. Compared with other algorithms, it is shown that our method compresses the model into a relatively smaller architecture with minor weights without any loss in accuracy for both the LeNet-300-100 and LeNet-5-Caffe. Particularly, our framework has the best compression performance in the middle layers of the network, which is 64 in the second layer. An important intuition about this phenomenon is that we introduce block sparse within each layer. Therefore, layers with more weights can induce a larger number of clusters, which can be further compressed. However, the model that is compressed by the proposed framework has more weights in the first layer compared to that of BC.

Then, the results shown in the WR column in table II suggest that most of the weights in the model shrink to zero after being compressed by our framework. On this item, our method is competitive with the other methods on LeNet-300-100 and maintains the best result, 0.6, on the LeNet5 architecture.

Especially on the error and CR indicators, our method has a high compression ratio while maintaining a reasonable accuracy. For LeNet-300-100, our method increases the compression ratio to 74 when the error rate starts to rise sharply at 1.88. Also, there is good balance between the compression rate and accuracy loss which is 713 and 0.89, respectively. This means that the proposed method can compress the model at a higher rate without significant loss in accuracy. Therefore, the compressed model can be utilized in mobile-enabled devices on the basis of ensuring both effectiveness and efficiency.

Fig. 3 presents the accuracy at different compression rates for two classical methods (blue and green curves) as well as our proposed method (the red curve). Compared with DC and SVD, in which accuracy begins to drop significantly when compression rate is below 3%, the accuracy of the network that is pruned by our method can maintain a 5% compression rate with no loss of accuracy. It even has 2% compression rate before beginning to dramatically decrease in accuracy.

TABLE II
EXPERIMENTAL RESULTS

| Architecture | Method | Pruned architecture (average bit precision) | error% | WR% | CR |
|---|---|---|---|---|---|
| LeNet 784-300-100 1.6 | SVD | 512-114-72 | 1.95 | 2.2 | 131 |
|  | BC | 311-86-14 | 1.9 | 10.6 | 59 |
|  | Proposed | 336-64-16 | 1.88 | 4.4 | 74 |
| LeNet5 20-50-800-500 0.9 | SVD | 14-19-242-131 | 0.91 | 0.7 | 349 |
|  | BC | 5-10-76-16 | 1.0 | 0.6 | 771 |
|  | Proposed | 12-10-45-12 | 0.89 | 0.6 | 713 |

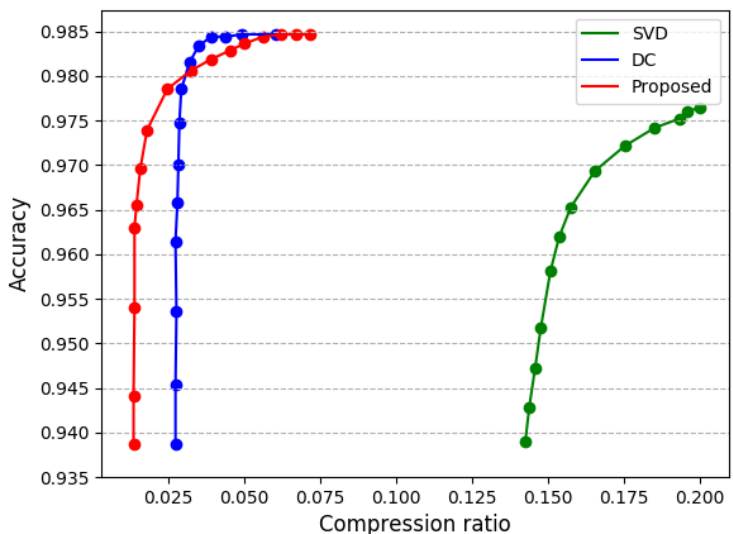

Fig. 3. Accuracy vs. Model size ratio after compression under different compression methods.

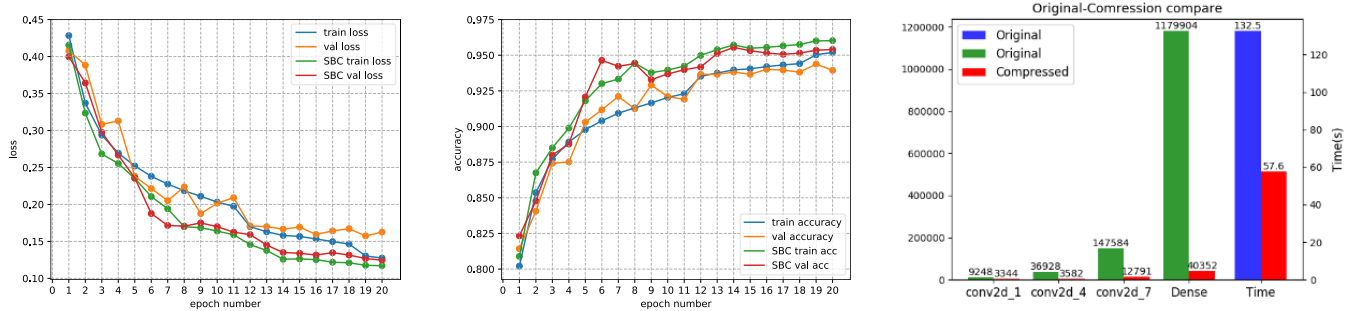

Fig. 4 From left to right, the loss curve, accuracy curve and number of parameters before and after compression by the proposed framework in typical layers are illustrated.

We present the experimental result on the HCD dataset in Fig. 4. To demonstrate the compression ability during the learning process of our model, the model is trained by traditional method and the proposed method on the same architecture. We can observe that the proposed method has a better convergence speed and higher accuracy. That is, in the 14th epoch, our network that is trained by the proposed method begins to converge with a 0.9523 accuracy rate. Furthermore, the bar shown in Fig. 4 shows that the effectiveness measured by the time-computing of the pruned model is significantly decreased by 2.3 ×. Also, the number of weights that are needed for the compressed model to obtain the same accuracy is significantly decreased. For the final dense layer, the compression rate is 29 ×. Therefore, these experimental results demonstrate that the compressed Cancer Detector can be utilized in the application of mobile-enabled devices.

## V. CONCLUSION

In this article, we discussed the model compression technologies which contribute to the utilization of a DNN-based model in mobile-enabled devices in a healthcare system and then proposed a novel structured Bayesian compression framework. As shown in the article, posteriors of weights in NN can shrink to zero with a large scale by introducing sparsity inducing priors to Bayesian learning methods. For the case of setting prior distribution, we exploit the idea of mixing sparsity inducing priors. We combine three different prior distributions that come from scale-mixtures of normals family, such that a flexibility and effectiveness prior can be inducted for further Bayesian inference. Then, to better utilize the structure of a deep neural network, we proposed the Structured Sparsity Learning scheme, which contains group sparse and block sparse to further improve the compression performance of our framework. With the help of proposed the Structured Bayesian compression framework, we can compress the DNN-based model with a high compression rate without a loss in accuracy, which is demonstrated by experiments on MNIST and the practical task Histopathologic Cancer Detection (HCD). As a conclusion, we expect that more mobile-enabled devices and real-time health services can be achieved by model compression methods, especially Bayesian compression. Also, because of the effectiveness and high performance of our proposed framework, we expect more works in this direction, introducing Bayesian learning with mixture prior and potential model structure utilization in Deep model compression.

## VI. CONCLUSION